\documentclass[journal]{IEEEtran}

\usepackage{xcolor,soul,framed} 

\colorlet{shadecolor}{yellow}
\usepackage[pdftex]{graphicx}
\graphicspath{{../pdf/}{../jpeg/}}
\DeclareGraphicsExtensions{.pdf,.jpeg,.png}

\usepackage[cmex10]{amsmath}
\usepackage{array}
\usepackage{mdwmath}
\usepackage{mdwtab}
\usepackage{eqparbox}
\usepackage{url}
\usepackage{amssymb}
\usepackage{graphicx}
\usepackage{booktabs}
\usepackage{multirow}

\begin{document}
    \title{Cross-Branch Orthogonality for Improved Generalization in Face Deepfake Detection}

\author{Tharindu~Fernando,~\IEEEmembership{Member,~IEEE,}       Clinton~Fookes,~\IEEEmembership{Senior Member,~IEEE,}
     
        Sridha~Sridharan,~\IEEEmembership{Life Senior Member,~IEEE,}
        ~and~ Simon~Denman,~\IEEEmembership{Member,~IEEE.}

\IEEEcompsocitemizethanks{\IEEEcompsocthanksitem T. Fernando, C. Fookes, S. Sridharan, and S.Denman are with The Signal Processing, Artificial Intelligence and Vision Technologies (SAIVT), Queensland University of Technology, Australia.\protect }}



\markboth{IEEE TRANSACTIONS ON IMAGE PROCESSING
}{Fernando \MakeLowercase{\textit{et al.}}: working title}

\maketitle

\begin{abstract}
Remarkable advancements in generative AI technology have given rise to a spectrum of novel deepfake categories with unprecedented leaps in their realism, and deepfakes are increasingly becoming a nuisance to law enforcement authorities and the general public. In particular, we observe alarming levels of confusion, deception, and loss of faith regarding multimedia content within society caused by face deepfakes, and existing deepfake detectors are struggling to keep up with the pace of improvements in deepfake generation. This is primarily due to their reliance on specific forgery artifacts, which limits their ability to generalise and detect novel deepfake types. To combat the spread of malicious face deepfakes, this paper proposes a new strategy that leverages coarse-to-fine spatial information, semantic information, and their interactions while ensuring feature distinctiveness and reducing the redundancy of the modelled features. A novel feature orthogonality-based disentanglement strategy is introduced to ensure branch-level and cross-branch feature disentanglement, which allows us to integrate multiple feature vectors without adding complexity to the feature space or compromising generalisation. Comprehensive experiments on three public benchmarks: FaceForensics++, Celeb-DF, and the Deepfake Detection Challenge (DFDC) show that these design choices enable the proposed approach to outperform current state-of-the-art methods by 5\% on the Celeb-DF dataset and 7\% on the DFDC dataset in a cross-dataset evaluation setting.

\end{abstract}


\IEEEpeerreviewmaketitle

\section{Introduction}

The fake video published by BuzzFeed showing an apparent speech by former US President Barack Obama that was in fact performed by Jordan Peele \cite{silverman2018spot} shows how easy it is to create convincing audio and video fakes. In recent years, we have seen an explosion of deep fakes, especially multimodal (video and audio) deep fakes. The extent and severe impact of fake multimedia content were clearly evident during the recent COVID-19 global pandemic \cite{balakrishnan2022infodemic} and the lead-up to the US federal 2020 election. Thus, the early detection of deep fakes is vital for stopping the spread of misinformation, which has influenced elections and led to serious consequences, including blackmail and fraud. 

To combat the surge of misleading deepfakes, a multitude of detection methods have emerged. However, there are significant concerns about whether these techniques can keep pace with the rapid advancements in deepfake generation \cite{yao2023towards, yin2024improving}. Specifically, recent studies have demonstrated that state-of-the-art (SOTA) deepfake detectors lack the ability to detect novel forgeries \cite{yao2023towards, zhang2024boosting}. Such generalisation is critical for detecting deepfakes, as it allows detectors to identify new types of manipulations, in particular those not present in the training data, thereby providing a safeguard against the constantly evolving deepfake generation landscape. Furthermore, generalised models could permit better abstraction and understanding of the broader concept of deepfakes, rather than being biased towards artifacts that are characteristics of individual methods \cite{lin2024preserving, zhang2024boosting, ba2024exposing}. In addition to providing more reliable and trustworthy decision-making, generalisation is important for eradicating unfair performance disparities across different demographic groups, preventing unfair targeting or exclusion \cite{lin2024preserving}.

\begin{figure} [!t]
    \centering
    \includegraphics[width=0.95\linewidth]{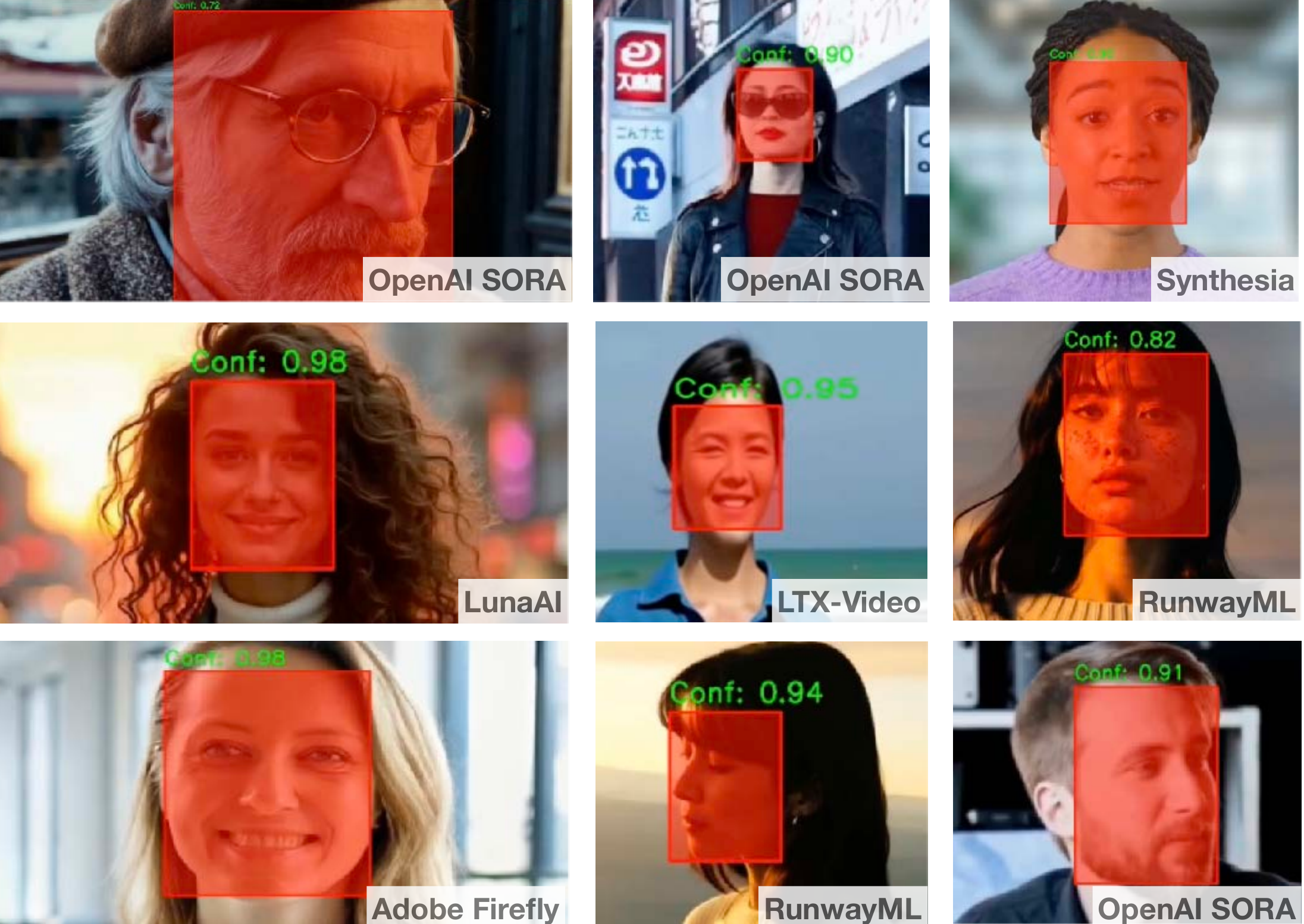}
    \caption{Fake faces identified by our Cross-Branch Orthogonal DeepFake Detection (CBO-DD) framework on completely unseen deepfake videos generated from the most recent generative AI video generation tools, including OpenAI SORA, RunwayML Gen-2, Adobe Firefly, LTX-Video, Synthesia, and Luma Dream Machine. The results demonstrate the generalisation of our model across different deepfake types as well as different ages, genders, ethnicities, and image characteristics.}
    \label{fig:hero_figure}
\end{figure}

Numerous recent works \cite{ding2024multi, coccomini2022combining, yasir2025lightweight, sun2021deepfakes} have demonstrated the utility of integrating features extracted from multiple pre-trained feature extractors at multiple scales, allowing models to effectively capture both local and global information and learn a more complete representation of the input by combining different feature types and scales. However, having multiple feature extractors and acquiring features at multiple scales could hinder a model's ability to generalise. Specifically, the feature extractors may capture redundant information, leading to overfitting \cite{ayinde2019regularizing}. Furthermore, integrating multiple feature extractors and multi-scale features increases the complexity of the feature space, interfering with the models' generalisation ability \cite{ hu2021model}. Therefore, a scheme is needed to diversify the feature selection process while obtaining informative cues across multiple feature branches. Moreover, the complexity of the feature space reduces the interpretability of interactions between features, making it harder for users to trust the model's predictions.

As a solution, we take inspiration from \cite{ahmed2024orco, li2020learning, wang2020orthogonal, li2019orthogonal} which discuss the importance of feature orthogonality when fusing multiple branches of information. In particular, orthogonal features minimize redundancy by ensuring that the model learns to capture diverse aspects of the data. This strengthens the model's generalisation ability and improves performance on unseen data \cite{ahmed2024orco, wang2020orthogonal}. Moreover, disentangled features are easier to interpret, as they are explicitly forced to represent distinct characteristics of the input \cite{li2020learning}.

Deviating from existing deepfake detection approaches, we implement a feature bottleneck via orthogonal disentanglement by projecting features into two lower-dimensional subspaces: (i) a \textit{Shared Component}, which captures common or overlapping information across distinct feature extraction branches; and (ii) a \textit{Disentangled Component}, which focuses on the unique, complementary aspects that are extracted from the specific branch. Moreover, extending beyond the current literature that considers single-level feature orthogonality, we demonstrate the effectiveness of hierarchical feature disentanglement in our proposed multi-branch architecture. Specifically, our framework enforces: (i) branch-level disentanglement, encouraging each branch to capture its own unique cues in the feature vectors extracted within the branch with some shared cross-branch representation; (ii) cross-branch disentanglement, enforcing an orthogonality constraint between branches to capture complementary aspects of the input. The result is an architecture that achieves unprecedented levels of generalisation across datasets and effectively detects completely unseen deepfake generation types (See Fig. \ref{fig:hero_figure}). 

Moreover, surpassing the most recent work by Ba et. al \cite{ba2024exposing}, which leverages Information Bottleneck (IB) theory to capture a compressed nonoverlapping representation of the input. The IB theory primarily focuses on data compression. Consequently, when applied to multiple feature streams, it may eliminate subtle yet crucial forgery clues that can be discovered across the feature streams. Deviated from this approach, we demonstrate how a diverse and complementary feature representation can be achieved by extending the concept of feature orthogonality to multiple branches. Additionally, we show how orthogonal disentanglement can be expanded to capture subtle cross-branch interactions.

The main novel technical contributions of this paper, in which we introduce the proposed Cross-Branch Orthogonal DeepFake Detection (CBO-DD) framework, can be summarised as follows:

\begin{enumerate}
\item We introduce a multi-branch architecture combining local spatial, global contextual, and emotional features for robust deepfake detection.
\item We propose a novel feature orthogonality-based disentanglement module that enforces both branch-level and cross-branch independence, enabling effective feature fusion without redundancy.
\item We show that our framework achieves strong generalisation across datasets, including unseen manipulations from state-of-the-art generative models, without any domain adaptation.
\item Our method outperforms existing approaches by up to 7\% in cross-dataset AUC, demonstrating state-of-the-art performance on FF++ \cite{rossler2019faceforensics++}, Celeb-DF \cite{li2020celeb}, and DFDC  \cite{dolhansky2020deepfake} benchmarks.
\end{enumerate}
\section{Related Work}

\subsection{Deepfake Detection}
The majority of the literature on deepfake detection has focused on the detection of artefacts left by the deepfake generation methods. For instance, in \cite{yang2019exposing}, the authors leverage artefacts in 3D head pose, which they identify based on inconsistencies in estimated 3D head pose when estimated from central and whole-of-face landmarks. This approach is motivated by the observation that face-swap technology only swaps faces in the central face region while keeping the outer contour of the face intact; hence, there exists a mismatch in the landmarks in fake faces. Similarly, movement of facial action units \cite{agarwal2019protecting}, eyebrows \cite{nguyen2020eyebrow}, and physiological measurements such as remote visual PhotoPlethysmoGraphy (PPG) have also been used in literature \cite{qi2020deeprhythm, ciftci2020fakecatcher} to identify artefacts. Another class of algorithms considers frequency domain artefacts that arise through compression errors or forgery. For example, the LipForensics \cite{haliassos2021lips} model considers the irregularities in the frequency of lip movement, whereas in \cite{frank2020leveraging, liu2021spatial} the authors suggest searching for ghost artifacts that arise due to an upsampling operation used in generative models. Despite the encouraging performance of artefact-based methods in within-dataset evaluation settings, these methods do not generalise well across different forgery categories, as they are tuned to detect only a handful of forgery clues. 

Multi-branch architectures have also been popular in deepfake detection. For instance, in \cite{kumar2020detecting}, five ResNet18 models have been used to extract local and global features. Moreover, the authors of \cite{rana2020deepfakestack} have leveraged pre-trained XceptionNet, MobileNet, ResNet101, InceptionV3, DenseNet121, InceptionReseNetV2, and DenseNet169 models as base learners in an ensemble of deepfake classifiers. Additionally, recent advancements in sequence learning techniques using transformer networks have led to numerous studies \cite{wodajo2021deepfake, zhao2023istvt} proposing the decomposition of spatial features extracted by CNNs into tokens. These tokens are then used with self-attention mechanisms to learn the relationships between them. However, none of these works have investigated the generalisation ability of the extracted features. When multiple feature extractors attend to the same spatial regions, they may extract redundant features. Moreover, self-attention-based dense modelling of the extracted features may increase the complexity of the feature space. While these features may exhibit superior classification performance by overfitting on dataset-specific artefacts, they fail to generalise well across different datasets.

\subsection{Generalisable Deepfake Detection}

There exist two popular methods within deepfake detection for achieving generalistaion: (i) supplementary data-based methods and (ii) domain adaptation-based methods. In supplementary data-based methods \cite{wang2022deepfake, zhao2021learning, shiohara2022detecting, Bai2023AUNetLR}, supplementary training data or self-supervised training objectives are used to provide additional information for training detectors and improving their generalisability. In contrast, domain adaptation-based methods \cite{tariq2021one, zhang2024boosting, kim2021fretal} transfer a detector trained in one domain to another (i.e., target domain) such that the detector can recognise the deepfakes in the target domain. While these methods demonstrate improved generalisation capabilities, their application in real-world conditions remains questionable. For instance, sourcing data for new deepfake generation types for re-training or domain adaptation can be challenging. Therefore, a framework that learns robust, generalisable, and non-redundant features from the training data without requiring re-training or domain adaptation is preferable.  

We note that a limited number of works have investigated the cross-distribution learning paradigms for improving the generalisation ability of deepfake detection methods. Considering such methods, \cite{yu2022improving} proposed a supervised common forgery tracing approach to learn to classify deepfakes across different datasets. In a different line of work, a hybrid approach is formulated in \cite{nadimpalli2022improving}, where the authors propose combining supervised learning and reinforcement learning to achieve better generalisation. Specifically, an RL agent is trained to select the top-k image augmentations for each test sample, which are most effective in distinguishing between real and fake images. The final classification (real or fake) is determined by averaging the CNN classification scores of all augmentations for each test image. Despite these advances, the generalisation ability of these methods is reliant upon the different real-world data distributions that the training data or augmentations could simulate. 

Most recently, in \cite{ba2024exposing}, the authors have proposed the use of Information Bottleneck (IB) theory for capturing a compressed, yet comprehensive feature representation for uncovering more forgery cues and improving the generalisation of deepfake detection. IB aims to find the best trade-off between accuracy and complexity, thereby extracting relevant forgery clues while discarding irrelevant information. Deviating from this approach, our work emphasises diverse and complementary feature extraction through orthogonal disentanglement, while facilitating cross-branch interactions through a shared latent space. In contrast, the authors of \cite{ba2024exposing} use multiple instances of the same pre-trained feature extractor to extract local features, which may limit the diversity and the comprehensiveness of the extracted features. Moreover, the direct extension of IB theory to multiple pre-trained feature extractor branches could hinder cross-branch interactions as information bottleneck theory is primarily focusing on compressing the data, potentially discarding subtle but crucial forgery clues that can be uncovered via the interactions between complementary feature streams. This can result in a model that is less sensitive to nuanced manipulations, reducing its effectiveness in detecting deepfakes. Therefore, the proposed method deviates from \cite{ba2024exposing} with respect to architectural choices and focuses on the areas of feature disentanglement as the theoretical foundation for improved generalisation.

\section{Methods}
In this section, we discuss our proposed approach. The main components that constitute our Cross-Branch Orthogonal DeepFake Detection (CBO-DD) framework are discussed in Sec. \ref{sec:overview}. The multi-branch encoder module that we use to extract multiple-scale and semantic abstractions of the input is introduced in Sec. \ref{sec:encoder}. Sec. \ref{sec:orthogonality} discusses the branch-level and cross-branch feature disentanglement strategy we implement to achieve better generalisation of the encoded features. In Sec. \ref{sec:classifier}, we present our pipeline for generating video-level deepfake classifications, and Sec. \ref{sec:losses} discusses the loss functions used for training the proposed CBO-DD architecture. Finally, implementation details of the framework are presented in Sec. \ref{sec:implementation}.

\subsection{Overview}\label{sec:overview}

\begin{figure*} [htbp]
    \centering
    \includegraphics[width=0.95\linewidth]{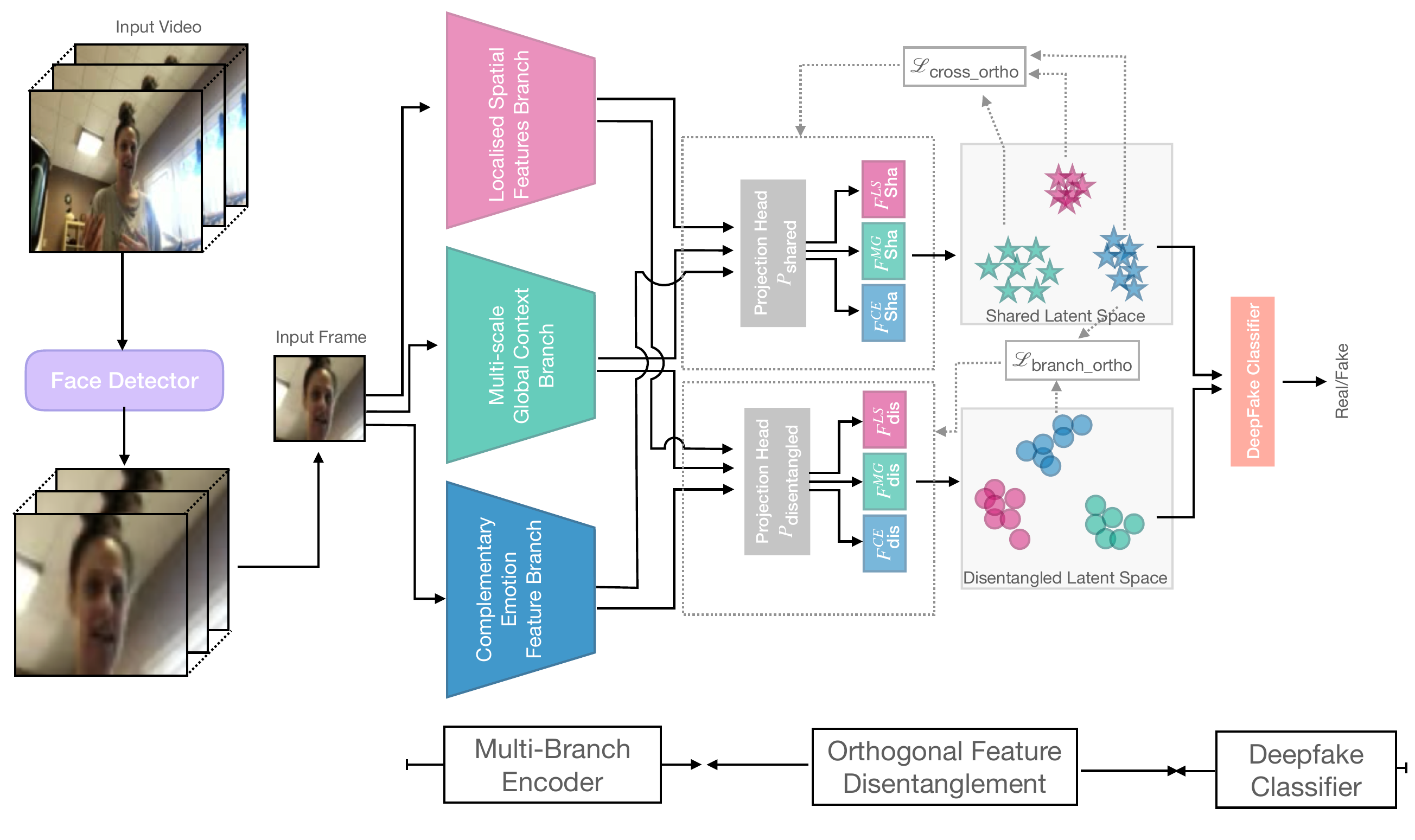}
    \caption{Method overview: We first extract frame-level facial bounding boxes from the input video. The multi-branch encoder module, which consists of a Localised Spatial Feature Branch, a Multi-scale Global Context Branch, and a Complementary Emotion Feature Branch, extracts multiple semantic features from the frame-level inputs. An Orthogonal Feature Disentanglement Module, which uses two projection heads, $P_{Shared}$ and $P_{disentangled}$, and branch-level $L_{\text{branch-ortho}}$ and cross-branch $L_{\text{cross-ortho}}$ Orthogonality losses, enforces branch-level and cross-branch feature disentanglement. Our deepfake classifier module utilises these disentangled features to generate frame-level classifications. Frame-level classifications are aggregated using a majority voting scheme to generate a video-level classification.}
    \label{fig:overview}
\end{figure*}

In this subsection, we provide an overview of the proposed CBO-DD framework. Our framework is composed of three main modules: a multi-branch encoder; an Orthogonal Feature Disentanglement Module that enforces branch-level and cross-branch feature disentanglement; and a deepfake classifier. These modules, the flow of information between them, and the objective functions used for their optimisation are illustrated in Fig. \ref{fig:overview}.

\subsection{Multi-Branch Encoder}\label{sec:encoder}
Our framework utilise three feature encoding branches to extract multiple semantic abstractions of the frame-level inputs. Our motivation is to adaptively capture diverse and complementary information from different aspects of the input frame, including localised spatial, multi-scale, and semantic clues. 

Specifically, as our frame-level feature extractors we use an EfficientNet \cite{tan2019efficientnet} pre-trained on the ImageNet dataset \cite{deng2009imagenet}, a Swin Transformer \cite{liu2021swin} which is also pre-trained on the ImageNet dataset, and HSEmotion \cite{savchenko2022classifying} -- a CNN-based emotion feature extraction model pre-trained on the Affectnet dataset \cite{mollahosseini2017affectnet}. EfficientNet excels at capturing local spatial details, such as edges and textures, which are essential for identifying fine-grained artifacts introduced during the deepfake generation process. On the other hand, Swin Transformer hierarchically captures global context and long-range dependencies through self-attention mechanisms, making it effective at identifying inconsistencies that span across different regions of the face. Numerous works have highlighted that fake faces often lack the emotional expression of a genuine face. As such, we incorporate the pre-trained HSEmotion feature extractor to complement our spatial feature extractors by capturing subtle emotional cues and discrepancies in facial expressions, further enhancing the model's ability to detect deepfakes. This combination of localised spatial representations, multi-scale global context awareness, and emotional analysis creates a robust and comprehensive feature representation, making our model effective at identifying a wide range of manipulations. Details of these 3 branches are presented in the following subsections.

\subsubsection{Localised Spatial Features Branch}

Formally, let $x_{\tau}$ denote a frame $\tau$ of the input video $v_{\eta}$ which is $T$ frames in length, where $v_{\eta} = [x_1, \ldots, x_{\tau}, x_T]$. Then our localised spatial feature extraction branch, which is formulated using the EfficientNet model, extracts the feature map, $F^{LS} \in \mathbb{R}^{\grave{C} \times \grave{H} \times \grave{W}}$, where $\grave{C}$ is the number of channels, and $\grave{H}$ and $\grave{W}$ are the height and width of the feature map, respectively, from the $l^{LS}$ layer of the EfficientNet architecture. Then, leveraging an Adaptive Average Pooling (AAP) layer, we segment the feature map, $F^{LS}$, into multiple non-overlapping segments, $F^{LS}_{seg} \in \mathbb{R}^{C \times \grave{k}_h \times \grave{k}_w}$, as: 
\begin{equation}
    F^{LS}_{seg} = \mathrm{AAP}(F^{LS}),
\end{equation}

where $\grave{k}_h$ and $\grave{k}_w$ are the height and width of the pooling window, and the operation of the AAP layer can be written as:
\begin{equation}
    F^{LS}_{\text{seg}}[\grave{C}, i, j] = \frac{1}{\grave{k}_h \grave{k}_w} \sum_{m=0}^{\grave{k}_h-1} \sum_{n=0}^{\grave{k}_w-1} F[\grave{C}, i \cdot s_h + m, j \cdot s_w + n],
    \label{eq:patch}
\end{equation}

where $s_h$ and $s_w$ are the strides in the height and width dimensions. Then, we flatten the $k_h$ and $k_w$ dimensions such that the extracted feature $F^{LS}_{seg}$ of the localised spatial feature branch is of shape $ C \times (k_h \times k_w)$.

\subsubsection{Multi-scale Global Context Branch}
In this branch, the frame $x_{\tau}$ is divided into non-overlapping windows of size $M \times M$ such that, $x_{\tau} \rightarrow \left\{ x_{{\tau},({i,j})} \right\}_{i,j=1}^{\frac{H}{M}, \frac{W}{M}}$. Then, self-attention is applied independently within each window. This allows the model to capture local context and interactions within each window. To capture cross-window interactions and global context in subsequent layers, the windows are shifted by a fixed number of pixels (e.g., half the window size), such that:
\begin{equation}
    \left\{ x_{{\tau},({i,j})} \right\}_{i,j=1}^{\frac{H}{M}, \frac{W}{M}} \rightarrow \left\{ x_{\tau, ({i + \frac{M}{2}, j + \frac{M}{2})}} \right\}_{i,j=1}^{\frac{H}{M}, \frac{W}{M}}.
    \label{eq:transformation}
\end{equation}
Using this structure, we hierarchically aggregate the local context to form the global context, progressively merging windows and increasing the receptive field. 

Let the feature maps extracted using this approach be denoted by $F^{MG}\in \mathbb{R}^{\dot{C} \times \dot{H} \times \dot{W}}$, where $\dot{C}$ is the number of channels, and $\dot{H}$ and $\dot{W}$ are the height and width of the feature map, respectively, from the $l^{MG}$ layer of the Swin Transformer architecture described above. Then, we apply the adaptive average pooling operation defined in Eq. \ref{eq:patch} across the spatial dimensions to compute the segmented feature map $F^{MG}_{seg} \in \mathbb{R}^{\dot{C} \times \dot{k}_h \times \dot{k}_w}$, and flatten the $\dot{k}_h$ and $\dot{k}_w$ dimensions such that the extracted feature $F^{MG}_{seg}$ of the multi-scale global context branch is of shape $ \dot{C} \times (\dot{k}_h \times \dot{k}_w)$.

\subsubsection{Complementary Emotion Feature Branch}
Similar to the previous branches, we extract a multi-dimensional feature map of shape $\breve{C} \times (\breve{k}_h \times \breve{k}_w)$ from a pre-trained HSEmotion model. Formally, let $F^{CE} \in \mathbb{R}^{\breve{C} \times \breve{H} \times \breve{W}}$ denote the output feature map of layer $l^{CE}$ of the HSEmotion model. Then, we apply Eq. \ref{eq:patch} across the spatial dimension to compute the segmented feature map $F^{CE}_{seg} \in \mathbb{R}^{\breve{C} \times \breve{k}_h \times \breve{k}_w}$, and flatten the $\breve{k}_h$ and $\breve{k}_w$ dimensions. 

\subsection{Modelling Relationships Between Features}

To model the relationships across the individual feature segments,  $ F^{\delta} \in [F^{LS}_{seg}, F^{MG}_{seg},$ and $F^{CE}_{seg}]$, we use a Multi-Head Self-Attention (MSA) mechanism. Specifically, each segment of a given branch is linearly projected to $D$ dimensional feature space, and the projected features are then passed through an MSA mechanism that helps in capturing relationships among the ${k}_h \times {k}_w$ segments. The resultant transformed feature, $F_\mathrm{trans} \in \mathbb{R}^{({k}^{\delta}_h \times {k}^{\delta}_w), D}$ then undergoes Global Average Pooling where we aggregate the information across the spatial dimension ${k}^{\delta}_h \times {k}^{\delta}_w$ as:
\begin{equation}
    F^{\delta}_{\text{pooled}} = \frac{1}{k^{\delta}_h \times k^{\delta}_w} \sum_{i=1}^{k^{\delta}_h \times k^{\delta}_w} F_{\text{trans}}[:, i, :]
\end{equation}

Using this approach, we obtain three feature vectors, $F^{LS}_{\text{pooled}}, F^{MG}_{\text{pooled}},$ and $ F^{CE}_{\text{pooled}}$, each representing the characteristics of the frame, $x_{\tau}$, with $D$ dimensions. To simplify the notation, in the subsequent sections we indicate the three feature vectors as $F^{LS}, F^{MG},$ and $ F^{CE}$.

\subsection{Orthogonal Feature Disentanglement Module}\label{sec:orthogonality}
In this section, we first describe the core concept behind the proposed Orthogonal Feature Disentanglement Module (OFDM) and then discuss how OFDM can be extended to enforce branch-level and cross-branch disentanglement. 

First, we split the feature representation into two components: (i) a \textit{Shared Component}, which captures shared or redundant information within the feature representation, and (ii) a \textit{Disentangled Component}, which captures the unique aspects within the feature vector. This is implemented using two projection matrices to project the input feature $F$ into two subspaces, $F_{\text{shared}}$ and $F_{\text{disentangled}}$ as:

\begin{equation}
\begin{aligned}
    F_{\text{shared}} &= P_{\text{shared}}(F), \\
    F_{\text{disentangled}} &= P_{\text{disentangled}}(F),
\end{aligned}
\end{equation}

where $P_{\text{shared}}$ and $ P_{\text{disentangled}}$ are learnable projection heads. Then, using a regularisation term, we enforce the two projected components to be orthogonal. Specifically, our objective is to minimise the squared Frobenius norm of the dot product between the two projections:

\begin{equation}
\begin{split}
    \min_{P_{\text{shared}}, P_{\text{disentangled}}} \mathcal{L}_{\text{ortho}} = & \min_{P_{\text{shared}}, P_{\text{disentangled}}} \left( \|P_{\text{shared}}(F)^T \right. \\
    & \left. P_{\text{disentangled}}(F)\|_F^2 \right).
\end{split}\label{eq:or}
\end{equation}

Next, we describe the process of implementing branch-level and cross-branch disentanglement, where we ensure that the features extracted from different branches are distinct and complementary, such that the branches can interact effectively with each other.

\subsubsection{Branch-Level Disentanglement}
We can directly extend OFDM to our 3 branches, where we use the $P_{\text{shared}}$ and $ P_{\text{disentangled}}$ projection heads to split the extracted features from each branch into shared and disentangled components. Formally, this can be written as:

\begin{equation}
\begin{aligned}
    F^{LS}_{\text{shared}} &= P_{\text{shared}}(F^{LS}), \\
    F^{LS}_{\text{disentangled}} &= P_{\text{disentangled}}(F^{LS}), \\ \\
    F^{MG}_{\text{shared}} &= P_{\text{shared}}(F^{MG}), \\
    F^{MG}_{\text{disentangled}} &= P_{\text{disentangled}}(F^{MG}), \\\\
    F^{CE}_{\text{shared}} &= P_{\text{shared}}(F^{CE}), \\
    F^{CE}_{\text{disentangled}} &= P_{\text{disentangled}}(F^{CE}).   
\end{aligned}
\end{equation}

Then, using the orthogonal loss formulation defined in Eq. \ref{eq:or}, we can define the branch-level disentanglement loss as:
  \begin{equation}
       \mathcal{L}_{\text{branch\_ortho}} = \sum_{\delta \in [LS, MG, CE]} \|P_{\text{shared}}(F^{\delta})^T \, P_{\text{disentangled}}(F^{\delta})\|_F^2,
\end{equation}
which ensures that the within-branch features are distinct and non-redundant.

\subsubsection{Cross-Branch Disentanglement}

In our OFDM, the shared components from each branch are projected into a common latent space, facilitating interactions between the branches. This enables effective fusion by leveraging the unique strengths of each branch and creates a more comprehensive representation such that the overall model captures complementary information from different feature streams. To ensure that the shared latent space contains only non-overlapping information, we compute the cross-branch orthogonality loss, which is implemented as the sum of the pairwise orthogonality losses between the shared components of different branches. This can be written as
\begin{equation}
    \mathcal{L}_{\text{cross\_ortho}} = \sum_{i,j\in [LS, MG, CE]} \left( \left\| F_{\text{shared}}^{(i)} \cdot F_{\text{shared}}^{(j)} \right\|_F^2 \right).
\end{equation}

\subsection{DeepFake Classifier}\label{sec:classifier}
To compute a comprehensive feature vector to represent the frame $x_{\tau}$, we concatenate the shared and disentangled feature vectors across the 3 branches to obtain the fused feature vector,
\begin{equation}
    F = [F^{LS}_{\text{shared}}; F^{LS}_{\text{disentangled}}; F^{MG}_{\text{shared}}; F^{MGE}_{\text{disentangled}}; F^{CE}_{\text{shared}}; F^{CE}_{\text{disentangled}}];
\end{equation}

where $[\cdot ; \cdot]$ denotes concatenation. As the features are disentangled and thus capturing diverse and complementary information, and the dimension of the projected features (i.e. $F^{\delta}_\text{shared}$, and $F^{\delta}_{\text{disentangled}}$) is significantly smaller than the original feature dimension, $D$, a simple concatenation based feature fusion is capable of generating a robust fused feature. Therefore, we employ a simple MLP layer as our classifier, which generates a binary classification denoting the authenticity of the input feature, $F$. Formally, let $F$ be the input feature vector, and $W$ and $b$ be the weights and bias of the MLP layer, respectively. The classifier can be represented as:
\begin{equation}
    \hat{y} = \sigma(WF + b)
\end{equation}
where $\hat{y}$ is the predicted probability of the input feature $F$ being fake, and $\sigma$ is the sigmoid activation function. Therefore, our CBO-DD framework analyses each frame, $x_{\tau}$, of the video, $v_{\eta}$, individually and predicts whether the frame is real or fake. Each frame's prediction is treated as a vote, and video-level classification is generated by considering the majority of the votes.   

\subsection{Loss Functions}\label{sec:losses} 
The overall loss function, $L$, which is used to train our CBO-DD framework is defined as follows,
\begin{equation}
    L = L_{\text{cls}} + \lambda_{\text{branch}} \cdot \mathcal{L}_{\text{branch\_ortho}} + \lambda_{\text{cross}} \cdot \mathcal{L}_{\text{cross\_ortho}} 
\end{equation}
where $\lambda_{\text{branch}}$ and $\lambda_{\text{cross}}$ are hyperparameters controlling the strength of the branch-level and cross-branch orthogonality constraints.

\subsection{Implementation Details}\label{sec:implementation}

Implementation of this framework is completed using PyTorch. The Adam \cite{kingma2014adam} optimiser with an initial learning rate of $1e^{-2}$, a decay of $1e^{-4}$, and a step size of 5 is used for optimisation. The model is trained for 100 epochs on an NVIDIA A100 GPU. The embedding size of the three pooled feature vectors, $F^{LS}, F^{MG}$, and $F^{CE}$, was experimentally chosen and was set to 2048.  Similarly, the dimensions of the projected features, $F_{\text{shared}}$, $F_{\text{disentangled}}$, $\lambda_{\text{branch}}$ and $\lambda_{\text{cross}}$ are set to $128$, $512$, $0.4$, and $0.25$, respectively. 

\section{Experiments}
In this section, we first introduce the details of the three public benchmarks that we used for our evaluations (Sec. \ref{sec:datasets}). The evaluation protocols, including evaluation metrics and settings, are presented in Sec. \ref{sec:evalprotocol}. The main experimental results where we compare our proposed method with existing state-of-the-art approaches are presented in Sec. \ref{sec:main_results}. Ablation evaluations that were conducted to demonstrate the effectiveness of the proposed innovations are provided in Sec. \ref{sec:ablations}. Finally, Sec. \ref{sec:time_complexity} discusses the time complexity of our CBO-DD model.

\subsection{Datasets}\label{sec:datasets}

Considering recent deepfake detection studies \cite{ba2024exposing,zhang2024boosting, chen2022self, zhao2021multi}, we conduct our evaluations using three public and large-scale deepfake detection benchmarks:(i) FaceForensics++ (FF++) \cite{rossler2019faceforensics++}, (ii) Celeb-DF \cite{li2020celeb}, and (iii) the Deepfake Detection Challenge (DFDC) \cite{dolhansky2020deepfake}. FF++ is one of the most widely used datasets in deepfake detection, offering 4000 fake videos generated from four different face manipulation methods: DeepFakes, Face2Face, FaceSwap, and NeuralTextures. There exist three compression levels in FF++, and data from compression level C23, which is the highest quality level, is used in our evaluations. Celeb-DF is another challenging benchmark in deepfake detection with forged faces with high visual realism. This dataset has two different versions, Celeb-DF-V1 and Celeb-DF-V2, and we used Celeb-DF-V2 in our experiments, which has 590 pristine and 5,639 manipulated videos. DFDC is one of the largest datasets designed for deepfake detection, consisting of more than 100,000 videos. The data has been sourced from 3,426 subjects, and fake faces have been produced with several Deepfake, GAN-based, and non-learned methods. As such, DFDC is considered one of the most challenging deepfake detection benchmarks.

\subsection{Evaluation Protocol}\label{sec:evalprotocol}

Following recent literature \cite{ba2024exposing, yin2024improving}, we report Area Under the Receiver Operating Characteristic Curve (AUC) as the evaluation metric. While classification accuracy is commonly used as the performance metric in classification tasks, it can be misleading in imbalanced datasets by favoring the majority class. In contrast, AUC remains immune to the distribution of the data as it considers the relative ranking of positive and negative samples. 

To better scrutinise the performance of the proposed CBO-DD model, we conduct experiments in both `within dataset' and `cross-dataset' evaluation settings. Under within-dataset protocol, we test the model's performance using unseen data from the same dataset it was trained on. This helps in understanding how well the model generalises to the same manipulation type; however, it does not reveal the model's robustness to unseen/new manipulation types. Therefore, we conduct an additional evaluation in a cross-dataset setting in which the model is tested on entirely different datasets that were not used during training, providing insights into the model's ability to generalise across diverse manipulation types. 

Since our framework can generate predictions at both frame and video levels, we report performance at both levels. 

\subsection{Comparisons with Existing State-of-the-art Methods}\label{sec:main_results}
In this section, we report results for the proposed model and compare with the existing state-of-the-art methods under within-dataset and cross-dataset evaluation settings. 

\textbf{Within Dataset Evaluations:} Tab. \ref{tab:within_dataset_comparison} provides the within-dataset comparisons, where we compare the performance of the proposed CBO-DD model with the most recent State-Of-The-Art (SOTA) methods. From these evaluations, it is clear that the CBO-DD model is capable of consistently outperforming existing SOTA methods across all considered benchmarks. For example, in the FF++ dataset, our method achieves a 1.2 \% improvement over the SOTA ResNet34 \cite{ba2024exposing} method, and in the DFDC dataset, we have outperformed the SOTA ResNet34 method by a significant 2.6 \%. These evaluations clearly exhibit the strengths of the proposed CBO-DD model in learning multiple complementary feature vectors that are representative of distinct facial forgeries in the datasets that it has been trained on. Moreover, by combining features explicitly trained to be orthogonal to each other, we avoid the need for complex feature fusion strategies such as cross-attention-based fusion \cite{coccomini2022combining} or specialised spatio-temporal feature extractors \cite{zhang2021detecting}, and our method can use a simple concatenation of the extracted features. Despite the simplicity, our CBO-DD model achieves a significant performance boost compared to these sophisticated architectures, illustrating the merits of our feature disentanglement strategy. 

\begin{table*}[htbp]
    \centering
    \renewcommand{\arraystretch}{1.2}
    \begin{tabular}{|l|c|l|c|l|c|}
        \hline
        \multicolumn{2}{|c|}{FF++(C23)} & \multicolumn{2}{c|}{Celeb-DF-V2} & \multicolumn{2}{c|}{DFDC} \\ 
        \hline
        Method & AUC$\uparrow$ & Method & AUC$\uparrow$ & Method & AUC$\uparrow$  \\ 
        \hline
        Xception \cite{rossler2019faceforensics} & 0.963 & DeepfakeUCL \cite{fung2021deepfakeucl} & 0.905 & Selim Seferbekov* & 0.882  \\ 
        Xception-ELA \cite{rossler2019faceforensics} & 0.948 & SBIs \cite{shiohara2022detecting} & 0.937 & NTechLab* & 0.880  \\ 
        SPSL \cite{masi2020two} & 0.943 & Agarwal et al. \cite{agarwal2020detecting} & 0.990 & Eighteen Years Old* & 0.886  \\ 
        Face X-ray \cite{li2020face} & 0.874 & Wu et al. \cite{wu2023generalizing} & 0.998 & WM* & 0.883  \\ 
        TD-3DCNN \cite{zhang2021detecting} & 0.722 & TD-3DCNN \cite{zhang2021detecting} & 0.888 & TD-3DCNN \cite{zhang2021detecting} & 0.790  \\ 
         Coccomini et al. \cite{coccomini2024mintime} & 0.913 & Coccomini et al. \cite{coccomini2024mintime} & 0.967 & Coccomini et al. \cite{coccomini2024mintime} & 0.951  \\
        F$^3$-Net \cite{qian2020thinking} & 0.981 & Xception \cite{rossler2019faceforensics} & 0.985 & Chugh et al. \cite{chugh2020not} & 0.907  \\ 
        FInfer \cite{hu2022finfer} & 0.957 & FInfer \cite{hu2022finfer} & 0.933 & FInfer \cite{hu2022finfer} & 0.829  \\ 
        Yin et al. \cite{yin2024improving} & 0.979 & Yin et al. \cite{yin2024improving} & - & Yin et al. \cite{yin2024improving} & -  \\
        ResNet34 \cite{ba2024exposing} & 0.983 & ResNet34 \cite{ba2024exposing} & \textbf{0.999} & ResNet34 \cite{ba2024exposing} & 0.939  \\ 
        \hline
        \textbf{CBO-DD} & \textbf{0.995} & \textbf{CBO-DD} & \textbf{0.999} & \textbf{CBO-DD} & \textbf{0.964}\\ 
        \hline
    \end{tabular}
    \caption{Within dataset evaluation results on FF++ \cite{rossler2019faceforensics++}, Celeb-DF-V2 \cite{li2020celeb}, and DFDC \cite{dolhansky2020deepfake} datasets, where we train and test the models using the same dataset. '*' denotes the top-4 teams in DFDC. Best results are shown in bold.}
    \label{tab:within_dataset_comparison}
\end{table*}

\textbf{Cross-Dataset Evaluations:} Tabs. \ref{tab:cross_dataset_comparisons} and \ref{tab:cross_dataset_comparisons2} provide the cross-dataset evaluations in terms of AUC at frame and video levels, respectively. These evaluations clearly exhibit the lack of generalisation that impacts the SOTA deepfake detectors, as they tend to overfit on the noisy artefacts in the training dataset rather than learning genealisable broader concepts of deepfakes. In contrast, our CBO-DD method has demonstrated superior cross-dataset generalization, outperforming all baseline methods in both datasets for both frame and video level evaluations. This intriguing capability results from the proposed branch-level and cross-branch level feature disentanglement strategy that ensures that learned features are non-redundant and non-overlapping, generating a compressed, yet comprehensive feature representation, bolstering generalisability. Moreover, our innovative cross-branch orthogonality formulation facilitates interactions between the branches, allowing our CBO-DD model to learn complex non-linear and complementary information. Specifically, our model achieves 4.5 \% and 8.7 \% performance gains at the frame level on the Celeb-DF-V2 and DFDC datasets, respectively, compared to the current SOTA method, ResNet34 \cite{ba2024exposing}. Similarly, at the video level, our CBO-DD model outperforms the current SOTA ResNet34 \cite{ba2024exposing} model by 5 \% and 9 \% on Celeb-DF-V2 and DFDC datasets, respectively. 

\begin{figure*}[htbp]
    \centering
    \includegraphics[width=.98\linewidth]{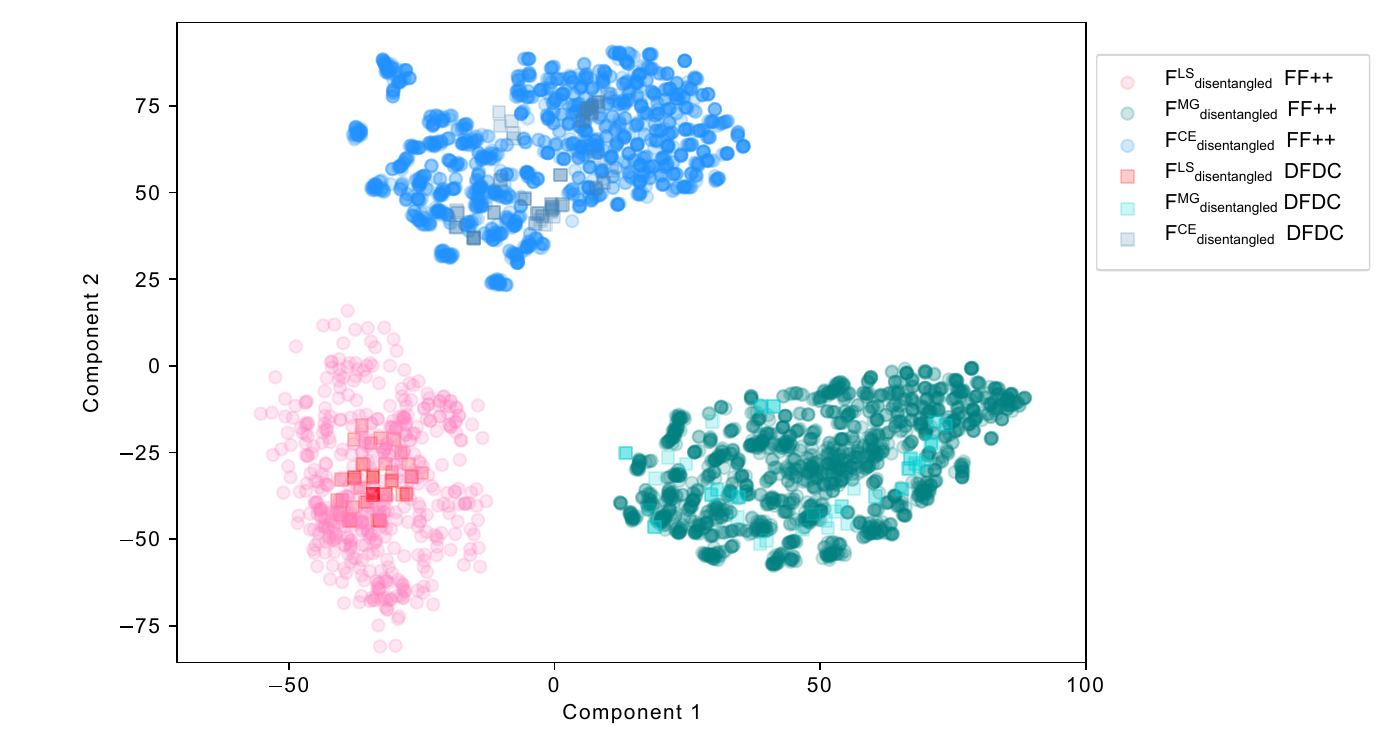}
    \caption{2D Visualisation of the distribution of the disentangled feature vectors ($F^{LS}_{\text{disentangled}}, F^{MG}_{\text{disentangled}}$, and $F^{CE}_{\text{disentangled}}$) in the cross-dataset evaluation. We indicate embeddings of the FF++ dataset training set as circles, and the DFDC dataset testing set as squares. }
    \label{fig:tsne}
\end{figure*}

To illustrate this superior genealisation capability of the proposed CBO-DD model we visualise the distribution of the  $F^{LS}_{\text{disentangled}}, F^{MG}_{\text{disentangled}}$, and $F^{CE}_{\text{disentangled}}$ embeddings. To plot the embeddings in 2 dimensions, we use t-SNE. In Fig. \ref{fig:tsne}, the disentangled embeddings of the FF++ training set are shown as circles, and the disentangled embeddings of the DFDC testing set are shown as squares. The plot shows distinct clusters for $F^{LS}_{\text{disentangled}}, F^{MG}_{\text{disentangled}}$, and $F^{CE}_{\text{disentangled}}$, indicating a clear disentanglement between the feature representations. Most importantly, the distribution of the testing embeddings fall in the latent space overlaps that of the training embeddings, $F^{LS}_{\text{disentangled}}, F^{MG}_{\text{disentangled}}$, and $F^{CE}_{\text{disentangled}}$, demonstrating the generalisability to the unseen DFDC testing samples. 

\begin{table*}[htbp]
    \centering
 \begin{tabular}{|l|c|c|c|}
        \hline
        \textbf{Method} & \textbf{Training dataset} & \textbf{Celeb-DF-V2} & \textbf{DFDC} \\
        \hline
        Xception \cite{rossler2019faceforensics} & FF++ & 0.778 & 0.636 \\
        DSP-FWA \cite{li2018exposing} & FF++ & 0.814 & - \\
        Meso4 \cite{afchar2018mesonet} & FF++ & 0.536 & - \\
        F$^3$-Net \cite{qian2020thinking} & FF++ & 0.712 & 0.646 \\
        Face X-ray \cite{li2020face} & PD & 0.742 & - \\
        Multi-Attention \cite{zhao2021multi} & FF++ & 0.674 & 0.680 \\
        Yin et al. \cite{yin2024improving} & FF++ & 0.705 & 0.674 \\
        RECCE \cite{cao2022end} & FF++ & 0.687 & 0.691 \\
        HCIL \cite{gu2022hierarchical} & FF++ & 0.790 & - \\
        LiSiam \cite{wang2022lisiam} & FF++ & 0.782 & - \\
        ICT \cite{dong2022protecting} & PD & 0.857 & - \\
        DCL \cite{sun2022dual} & FF++ & 0.823 & - \\
        IID \cite{huang2023implicit} & FF++ & 0.838 & - \\    
        ResNet34 \cite{ba2024exposing} & FF++ & 0.864 & 0.721 \\
        \hline
        \textbf{CBO-DD} & FF++ & \textbf{0.903} & \textbf{0.784} \\
        \hline
    \end{tabular}
    \caption{Cross-dataset frame-level AUC results on the Celeb-DF-V2 \cite{li2020celeb} and DFDC \cite{dolhansky2020deepfake} datasets. We train the models using the FF++ \cite{rossler2019faceforensics++} dataset. 'PD' denotes private data. Best results are shown in bold.}
    \label{tab:cross_dataset_comparisons}
\end{table*}

\begin{table*}[htbp]
    \centering
    \begin{tabular}{|l|c|c|c|}
        \hline
        \textbf{Method} & \textbf{Training dataset} & \textbf{Celeb-DF-V2} & \textbf{DFDC} \\
        \hline
        Xception \cite{rossler2019faceforensics} & FF++ & 0.737 & 0.709 \\
        F$^3$-Net \cite{qian2020thinking} & FF++ & 0.757 & 0.709 \\
        PCL+I2G \cite{zhao2021learning} & PD & 0.900 & 0.675 \\
        FST-Matching \cite{dong2022explaining} & FF++ & 0.894 & - \\
        LipForensics \cite{haliassos2021lips} & FF++ & 0.824 & 0.735 \\
        FTCN \cite{zheng2021exploring} & FF++ & 0.869 & 0.710 \\
        Luo et al. \cite{luo2021generalizing} & FF++ & - & 0.797 \\
        ResNet-34+ SBIs \cite{shiohara2022detecting} & PD & 0.870 & 0.664 \\
        EFNB4+ SBIs \cite{shiohara2022detecting} & PD & 0.932 & 0.724 \\
        RATF \cite{gu2022region} & FF++ & 0.765 & - \\
        Li et al. \cite{liwavelet} & FF++ & 0.848 & - \\
        AltFreezing \cite{Wang2023AltFreezingFM} & FF++ & 0.895 & - \\
        AUNet \cite{Bai2023AUNetLR} & PD & 0.928 & 0.738 \\
        ResNet34 \cite{ba2024exposing} & FF++ & 0.936 & 0.754 \\
        \hline
        \textbf{CBO-DD} & FF++ & \textbf{0.979} & \textbf{0.822} \\
        \hline
    \end{tabular}
        \caption{Cross-dataset video-level AUC results on Celeb-DF-V2 \cite{li2020celeb} and DFDC \cite{dolhansky2020deepfake} datasets. We train the models using the FF++ \cite{rossler2019faceforensics++} dataset. 'PD' denotes private data. Best results are shown in bold.}
    \label{tab:cross_dataset_comparisons2}
\end{table*}

\begin{figure*}[htbp]
    \centering
    \includegraphics[width=.98\linewidth]{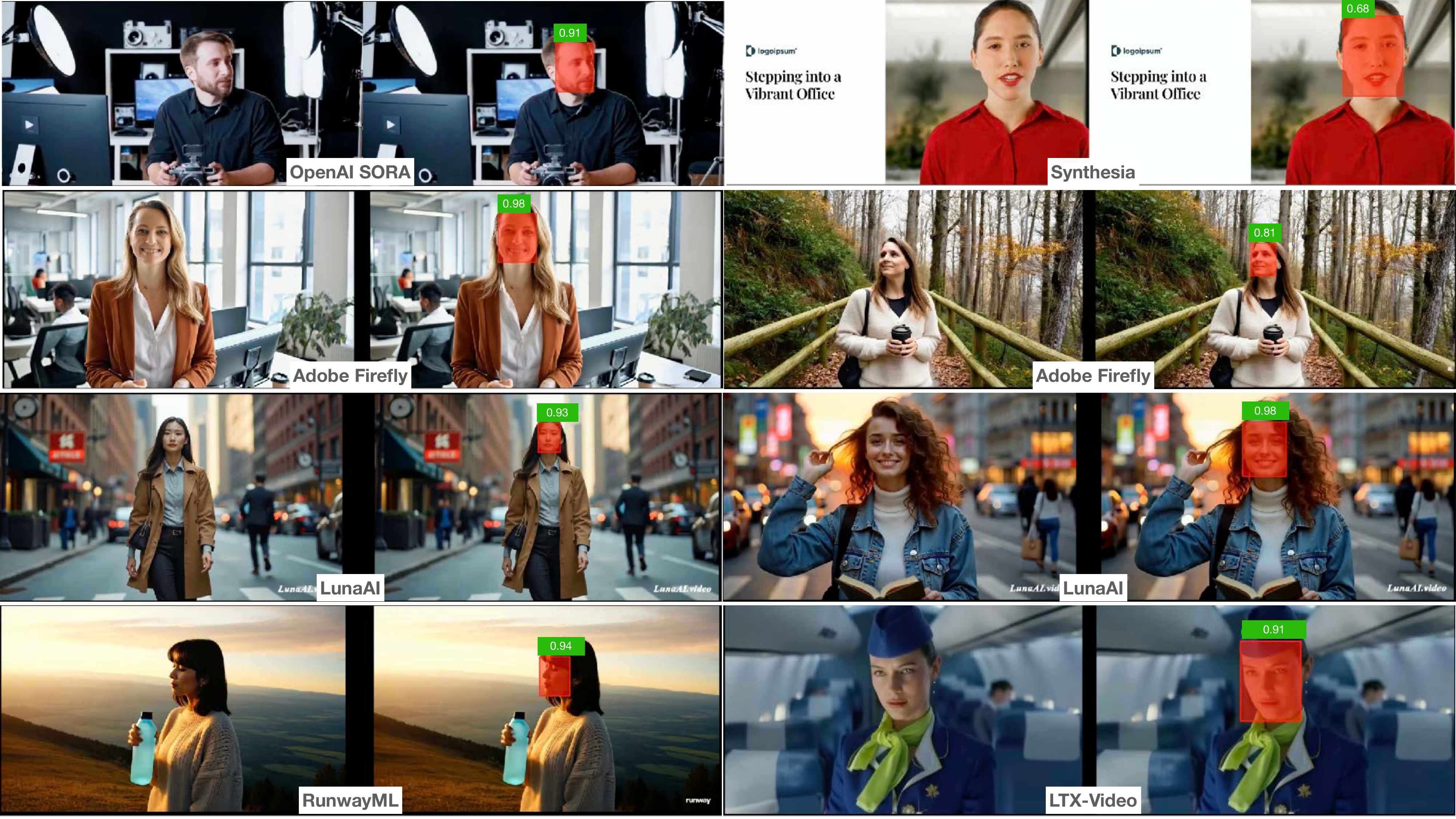}
    \caption{Qualitative results of our CBO-DD model (trained on FF++ dataset) when tested on videos generated by completely unseen GenAI video generation tools (OpenAI SORA, RunwayML Gen-2, Adobe Firefly, LTX-Video, Synthesia, and Luma Dream Machine). We visualise the sample frames from the videos showing the detected face, along with the video-level deepfake detection confidence, which has been generated by averaging the frame-level detection confidence. For additional visualisations, please refer to the supplementary material.}
    \label{fig:genAI}
\end{figure*}

To further illustrate the superior generalisation capabilities of the proposed method, we conduct an additional evaluation by generating deepfake videos using the most recent generative AI (GenAI) video generation tools, including OpenAI SORA \footnote{https://openai.com/sora/}, RunwayML Gen-2 \footnote{https://app.runwayml.com/login}, Adobe Firefly \footnote{https://www.adobe.com/au/products/firefly.html}, LTX-Video \footnote{https://www.lightricks.com/}, Synthesia \footnote{https://www.synthesia.io/} and Luma Dream Machine \footnote{https://lumalabs.ai/dream-machine}, and testing the CBO-DD model trained on the FF++ dataset on these videos. It should be noted that the CBO-DD model has never seen these manipulations during training. Fig. \ref{fig:genAI} provides qualitative visualisations of our model in which we have indicated the video-level confidence of the CBO-DD model that the video is fake. This has been generated by averaging the frame-level deepfake detection confidence. As expected, our model has been able to detect manipulations generated by these most recent GenAI video generation tools, demonstrating that the proposed approach offers a robust, non-redundant, comprehensive, and complementary deepfake feature learner to keep up with the constantly evolving deepfake generation technology. For additional visualisations, please refer to the supplementary material.

\subsection{Ablation Evaluations}\label{sec:ablations}
We hypothesize that our design choices, (i) the proposed multi-branch architecture that captures distinct and complementary features from the input, (ii) the proposed branch-level orthogonal feature disentanglement, and (iii) our innovative cross-branch orthogonal feature disentanglement, collectively contribute to the robustness of our model. Therefore, we conducted a series of ablation studies systematically analysing the impact of these individual innovations. For a complete analysis, all ablation experiments were conducted using both within-dataset and cross-dataset protocols at the frame level. In the within-dataset setting, we train the ablation models using the training set of the FF++ dataset and test the models using the validation set of the FF++ dataset. In the cross-dataset setting, we train the ablation models using the training set of the FF++ dataset and test the models using the validation set of the DFDC dataset. 

\subsubsection{Effects of Multi-Branch Architecture}
We study the effects of our multi-branch architecture by generating six ablation variants of the proposed CBO-DD model: (i) BO - w/o [MG, CE]: a model with only the localised spatial features branch and with only branch-level disentanglement; (ii) BO - w/o [LS, CE]: a model with only the multi-scale global context branch and with only branch-level disentanglement; (iii) BO - w/o [LS, MG]: a model with only the complementary emotion feature branch and with only branch-level disentanglement; (iv) CBO - w/o [LS]: the proposed model without the localised spatial features branch; (v) CBO - w/o [MG]: the proposed model without the multi-scale global context branch; and (vi) CBO - w/o [CE]: the proposed model without the complementary emotion feature branch.

\begin{table}[htbp]
  \centering
  \resizebox{.9\linewidth}{!}{%
\begin{tabular}{|l|c|cc|}
\hline
\multirow{2}{*}{Method} & \multirow{2}{*}{Trained On} & \multicolumn{2}{c|}{Tested On}   \\ \cline{3-4} 
                        &                             & \multicolumn{1}{c|}{FF++} & DFDC \\ \hline
BO - w/o {[}MG, CE{]}   & FF++                        & \multicolumn{1}{c|}{0.795}     &  0.623    \\ \hline
BO - w/o {[}LS, CE{]}   & FF++                        & \multicolumn{1}{c|}{0.823}     &  0.647    \\ \hline
BO - w/o {[}LS, MG{]}   & FF++                        & \multicolumn{1}{c|}{0.852}     & 0.698     \\ \hline
CBO - w/o {[}LS{]}      & FF++                        & \multicolumn{1}{c|}{0.990}     &   0.735   \\ \hline
CBO - w/o {[}MG{]}      & FF++                        & \multicolumn{1}{c|}{0.985 }      &    0.727  \\ \hline
CBO - w/o {[}CE{]}      & FF++                        & \multicolumn{1}{c|}{0.976}       &   0.719   \\ \hline
\textbf{CBO-DD }                 & FF++                        & \multicolumn{1}{c|}{\textbf{0.994}}     &  \textbf{0.787}    \\ \hline
\end{tabular}}
\caption{Effect of the Proposed Multi-Branch Architecture. The best results are highlighted in bold.}
    \label{tab:multi_branch_ablation}
\end{table}

The results of this comparison are shown in Tab. \ref{tab:multi_branch_ablation}. We observe significant contributions from all 3 branches of our CBO-DD model. In particular, we observe that the multi-scale global context and complementary emotion branches play pivotal roles in both within-dataset and cross-dataset evaluation settings, demonstrating the utility of our multi-branch architecture for improving robustness and enhancing generalisation of the CBO-DD model. 

\begin{figure*}[htbp]
    \centering
    \includegraphics[width=.8\linewidth]{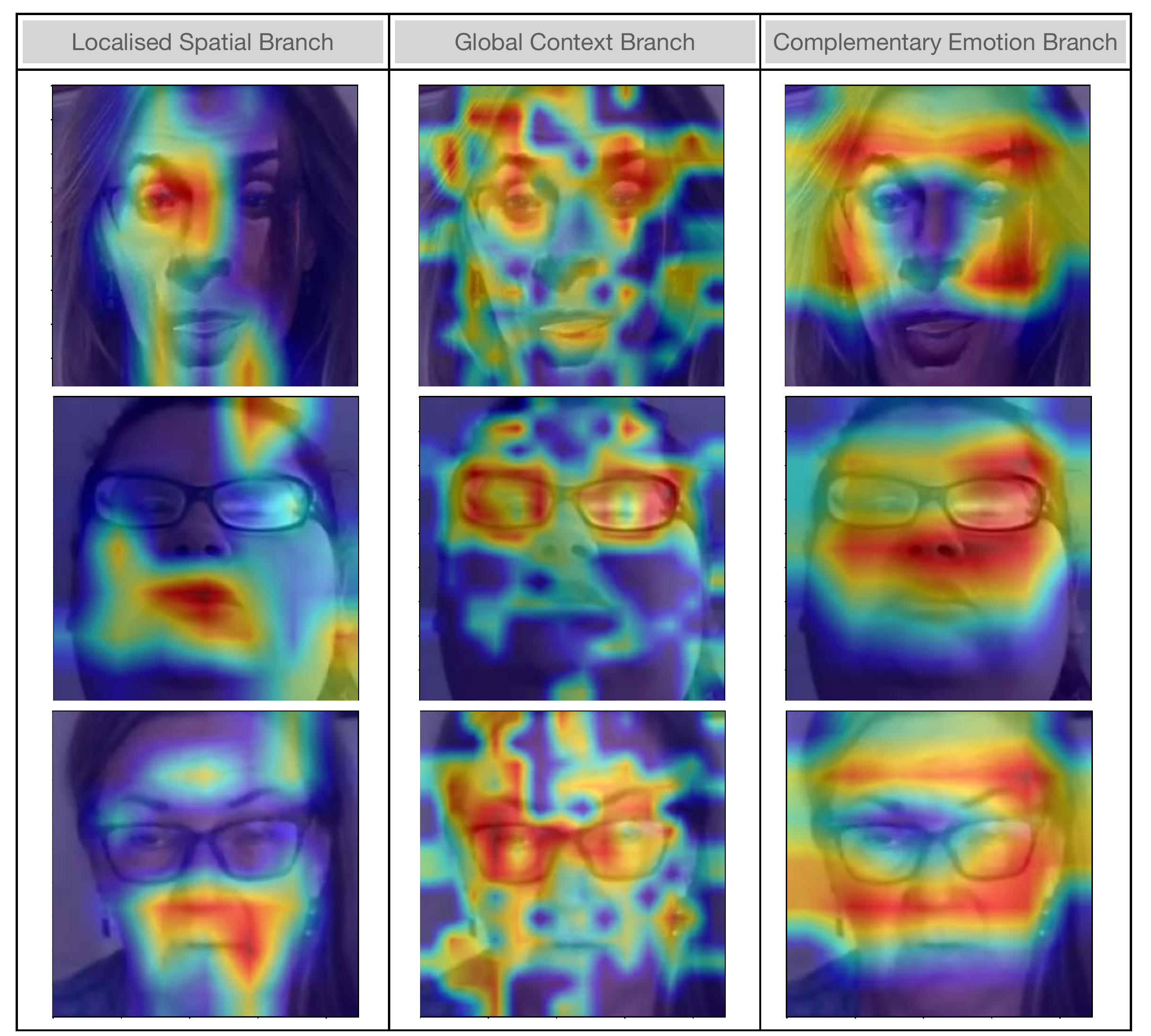}
    \caption{Visualisation of feature saliency maps derived from three branches, Localised Spatial Branch ($LS$), Multiscale Global ($MG$) context branch, and Complementary Emotion Branch ($CE$), for sample video frames in the DFDC test dataset.}
    \label{fig:gradcam}
\end{figure*}

In addition to quantitative results, in Fig. \ref{fig:gradcam}, we visualise the feature saliency maps extracted from three branches for sample frames of the test set from the DFDC dataset. These visualisations clearly illustrate that different branches have attended to different regions in the input frame. Moreover, we are able to see that multiple spatial regions in the input have been aggregated when extracting the global context of the input in the $MG$ branch, while in the $LS$ branch, local input-specific regions have been attended to. The $CE$ branch has generated complementary emotion-specific features via referring to the other salient regions in the face that provide emotion-related information. Furthermore, we observe that the spatial regions attended by the 3 regions are generally non-overlapping. 

\subsubsection{Effects of Branch-Level and Cross-Branch Orthogonal Feature Disentanglement}
In this experiment, we evaluate the effectiveness of the proposed branch-level and cross-branch orthogonal feature disentanglement processes. To evaluate these, we generated three ablation variants of the proposed model: (i) MB - w/o [BO, CBO]: a multi-branch model without branch-level and cross-branch orthogonal feature disentanglement; (ii) MB - w/o [CBO]: a multi-branch model without cross-branch orthogonal feature disentanglement; and (iii) MB - w/o [BO]: a multi-branch model without branch-level orthogonal feature disentanglement. 

\begin{table}[htbp]
    \centering
    \resizebox{.9\linewidth}{!}{%
    \begin{tabular}{|l|c|cc|}
\hline
\multirow{2}{*}{Method} & \multirow{2}{*}{Trained On} & \multicolumn{2}{c|}{Tested On}   \\ \cline{3-4} 
                        &                             & \multicolumn{1}{c|}{FF++} & DFDC \\ \hline
MB - w/o [BO, CBO]   & FF++                        & \multicolumn{1}{c|}{0.780}     & 0.612     \\ \hline
MB - w/o [CBO]   & FF++                        & \multicolumn{1}{c|}{0.964}     &   0.732   \\ \hline
MB - w/o [BO]   & FF++                        & \multicolumn{1}{c|}{0.941}     &  0.720   \\ \hline
CBO-DD                  & FF++                        & \multicolumn{1}{c|}{\textbf{0.994}}     &   \textbf{0.787}   \\ \hline
\end{tabular}}
    \caption{Effect of the Proposed Branch-Level and Cross-Branch Orthogonal Feature Disentanglement. The best results are highlighted in bold.}
    \label{tab:disentanglement_ablation}
\end{table}

From the results in Tab. \ref{tab:disentanglement_ablation}, we can confirm the utility of implementing both branch-level and cross-branch disentanglement.  We refer the reader to the rows corresponding to MB - w/o [CBO]  and MB - w/o [BO]  in Tab. \ref{tab:disentanglement_ablation}, where we observe a significant performance gain compared to the MB - w/o [BO, CBO] model which does not contain branch-level and cross-branch feature disentanglement. Most importantly, we observe some improvement in the cross-dataset generalisation in the model when at least one of these schemes is implemented, however, a further significant performance gain is achieved by the proposed CBO-DD model, which incorporates both branch-level and cross-branch feature disentanglement. This is because the branch-level and cross-branch feature disentanglement schemes collectively enable both shared and disentangled features across the branches to be non-overlapping and complementary, allowing us to generate a highly representative fused feature vector through simple concatenation. This simplifies classification and improves separation between real and fake samples. Therefore, using this experiment, we can confirm the necessity of both branch-level and cross-branch feature disentanglement schemes within our framework. 

\subsection{Time Complexity}\label{sec:time_complexity} We conduct a comprehensive time complexity analysis using four state-of-the-art fake deepfake detection models, including \cite{coccomini2022combining}, Xception \cite{rossler2019faceforensics}, and \cite{ba2024exposing}. These methods are chosen based on the public availability of their implementations. We measure the average time taken to generate 100 video level classifications, including the time taken for pre-processing the input and feature extraction, using a single NVIDIA A100 GPU. The evaluation results are presented in Tab. \ref{tab:time_complexity}. These results illustrate that the proposed CBO-DD model has been able to achieve substantial performance gains compared to these SOTA models without sacrificing its efficiency.

\begin{table}[htbp]

\label{tab:time_complexity}
\centering
\begin{tabular}{|c|c|c|}
\hline
Model    & Total Params (M) & Runtime (in Sec) \\ \hline
Xception \cite{rossler2019faceforensics}   &     23          &  0.82          \\ \hline
ResNet34 \cite{ba2024exposing}   &    87           &   2.89         \\ \hline
Coccomini et al. \cite{coccomini2024mintime}  &   101           &   3.19        \\ \hline
CBO-DD (Ours)  &  104           &   3.87          \\ \hline
\end{tabular}
\caption{Time Complexity Analysis: The parameter count in millions and the time taken to generate 100 video level classifications using a single NVIDIA A100 GPU}
\end{table}

\section{Current Limitations and Future Directions}
We observe two limitations of CBO-DD: (i) Our feature orthogonality-based feature disentanglement strategy is a purely data-driven approach and does not incorporate any prior knowledge regarding the features or their significance. However, if prior knowledge is available, it can be utilised for the configuration of the orthogonal feature disentanglement module, which could improve the convergence and yield more robust disentanglement. Furthermore, prior knowledge can be incorporated as guided supervision signals or regularisation constraints, which could help in separating the underlying factors of variation more effectively. Future research efforts could be directed to designing a hybrid approach where prior knowledge regarding the features is combined with feature orthogonality to learn a comprehensive feature representation. (ii) While our evaluations (Tab. \ref{tab:time_complexity}) show that the proposed model has comparable computational complexity to existing state-of-the-art deepfake detection models, it has not been tested on edge devices like smartphones. Despite enhancing deepfake detection robustness with complementary features, the multi-branch architecture's use of computationally expensive transformer-based backbones increases computational cost. Therefore, it may not be suitable for edge deployment. Future research could explore model pruning or distillation strategies to improve the efficiency of the CBO-DD model.

\section{Conclusion}
This paper presented a Cross-Branch Orthogonal DeepFake Detection (CBO-DD) framework for accurate detection of face deepfakes. One of our primary aims is to achieve cross-dataset generalisation without the need for laborious fine-tuning or domain adaptation. Our proposed multi-branch architecture, combined with a feature orthogonality-based disentanglement strategy, captures highly discriminative and complementary features. This approach provides a comprehensive view of deepfakes, avoiding overfitting to dataset-specific artifacts and achieving unprecedented levels of generalisation. Extensive experiments were conducted on three public benchmarks: FaceForensics++, Celeb-
DF and the Deepfake Detection Challenge (DFDC), which demonstrated the ability of the proposed framework to outperform the current state-of-the-art algorithms by significant margins.

\section*{Acknowledgment}
The research was supported by the Australian Government through the Office of National Intelligence Postdoctoral Grant awarded to the primary author under Project NIPG-2024-022.

\bibliographystyle{IEEEtran}
\bibliography{IEEEabrv,Bibliography}

\begin{IEEEbiography}[{\includegraphics[width=1in,height=1.25in,clip,keepaspectratio]{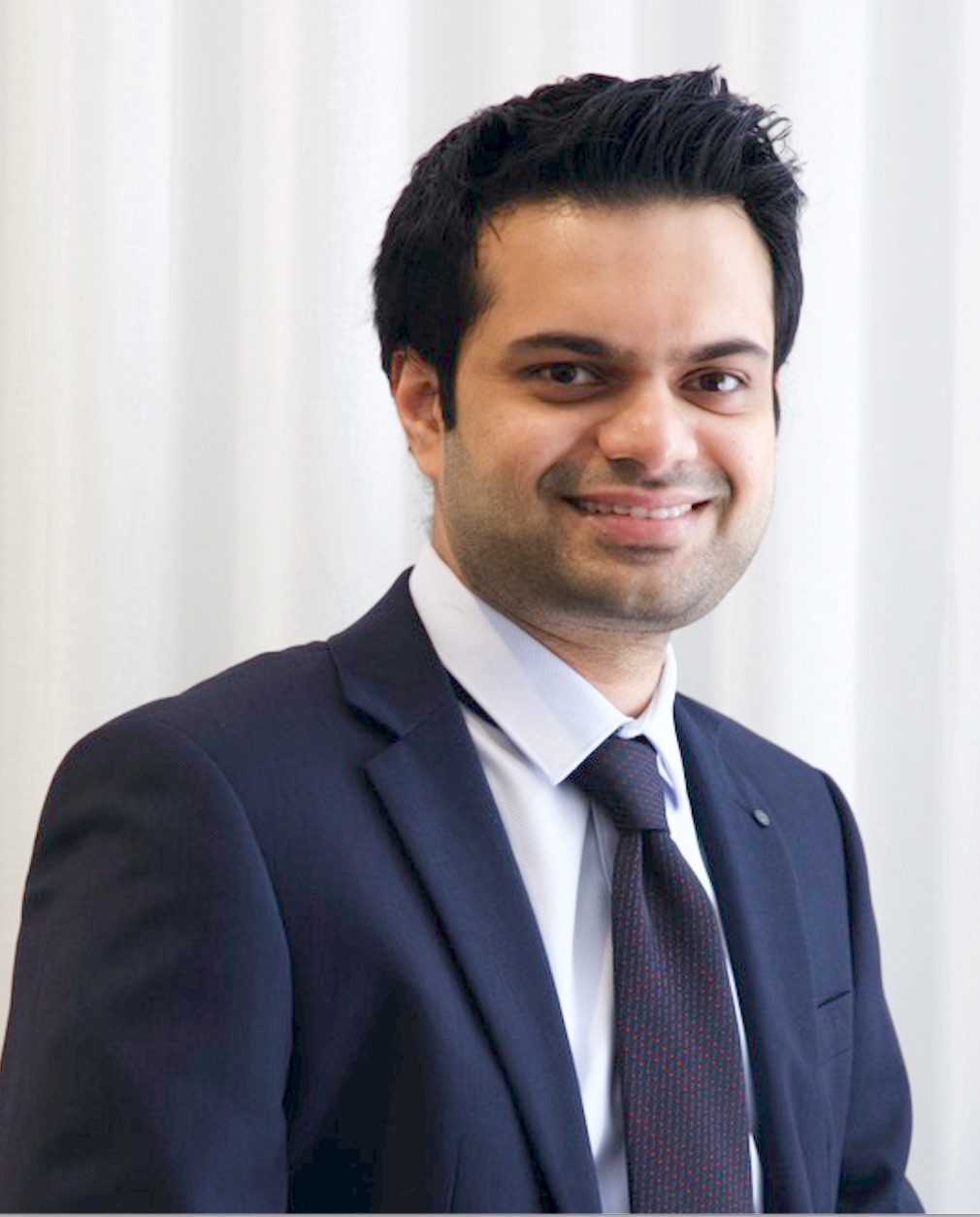}}]{Tharindu Fernando } received his BSc (special degree in computer science) from the University of Peradeniya, Sri Lanka, and his PhD from Queensland University of Technology (QUT), Australia. He is currently a Postdoctoral Research Fellow in the Signal Processing, Artificial Intelligence, and Vision Technologies (SAIVT) research program at the School of Electrical Engineering and Robotics at Queensland University of Technology (QUT). He is a recipient of the 2019 QUT University Award for Outstanding Doctoral Thesis, the QUT Early Career Researcher Award in 2022, the QUT Faculty of Engineering Early Career Achievement Award in 2024, and the 2024 National Intelligence Post-Doctoral Grant. His research interests include Artificial Intelligence, Computer Vision, Deep Learning, Bio Signal Processing, and Video Analytics.
\end{IEEEbiography}

\begin{IEEEbiography}[{\includegraphics[width=1in,height=1.25in,clip,keepaspectratio]{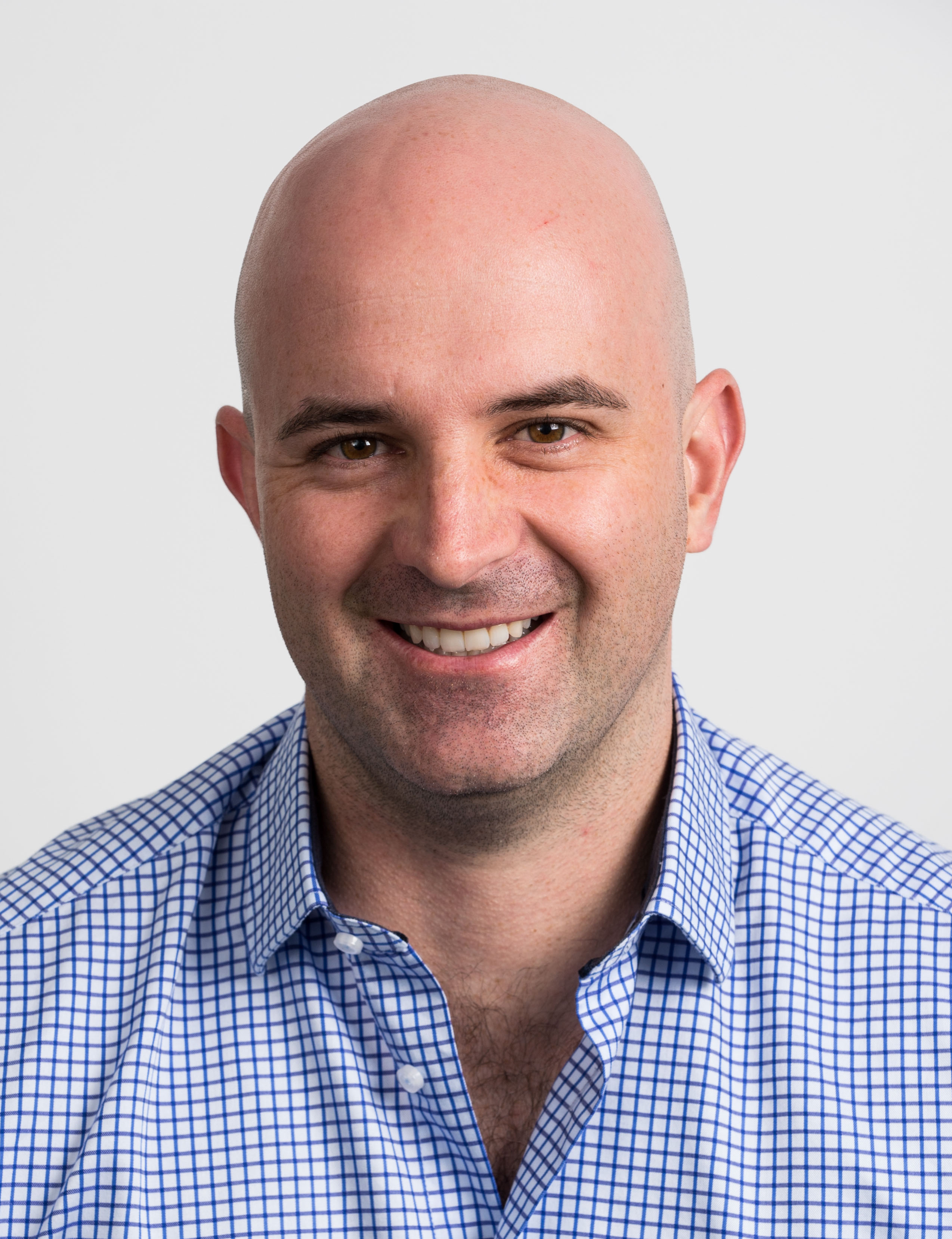}}]{Clinton Fookes}{\space}(Senior Member, IEEE) received the B.Eng. in Aerospace/Avionics, the MBA degree, and the Ph.D. degree in computer vision. He is currently the Associate Dean Research, a Professor of Vision and Signal Processing, and Co-Director of the SAIVT Lab (Signal Processing, Artificial Intelligence and Vision Technologies) with the Faculty of Engineering at the Queensland University of Technology, Brisbane, Australia. His research interests include computer vision, machine learning, signal processing, and artificial intelligence. He serves on the editorial boards for IEEE TRANSACTIONS ON IMAGE PROCESSING and Pattern Recognition. He has previously served on the Editorial Board for IEEE TRANSACTIONS ON INFORMATION FORENSICS AND SECURITY. He is a Fellow of the International Association of Pattern Recognition, a Fellow of the Australian Academy of Technological Sciences and Engineering, and a Fellow of the Asia-Pacific Artificial Intelligence Association. He is a Senior Member of the IEEE and a multi-award winning researcher including an Australian Institute of Policy and Science Young Tall Poppy, an Australian Museum Eureka Prize winner, Engineers Australia Engineering Excellence Award, Australian Defence Scientist of the Year, and a Senior Fulbright Scholar.
\end{IEEEbiography}

\begin{IEEEbiography}[{\includegraphics[width=1in,height=1.25in,clip,keepaspectratio]{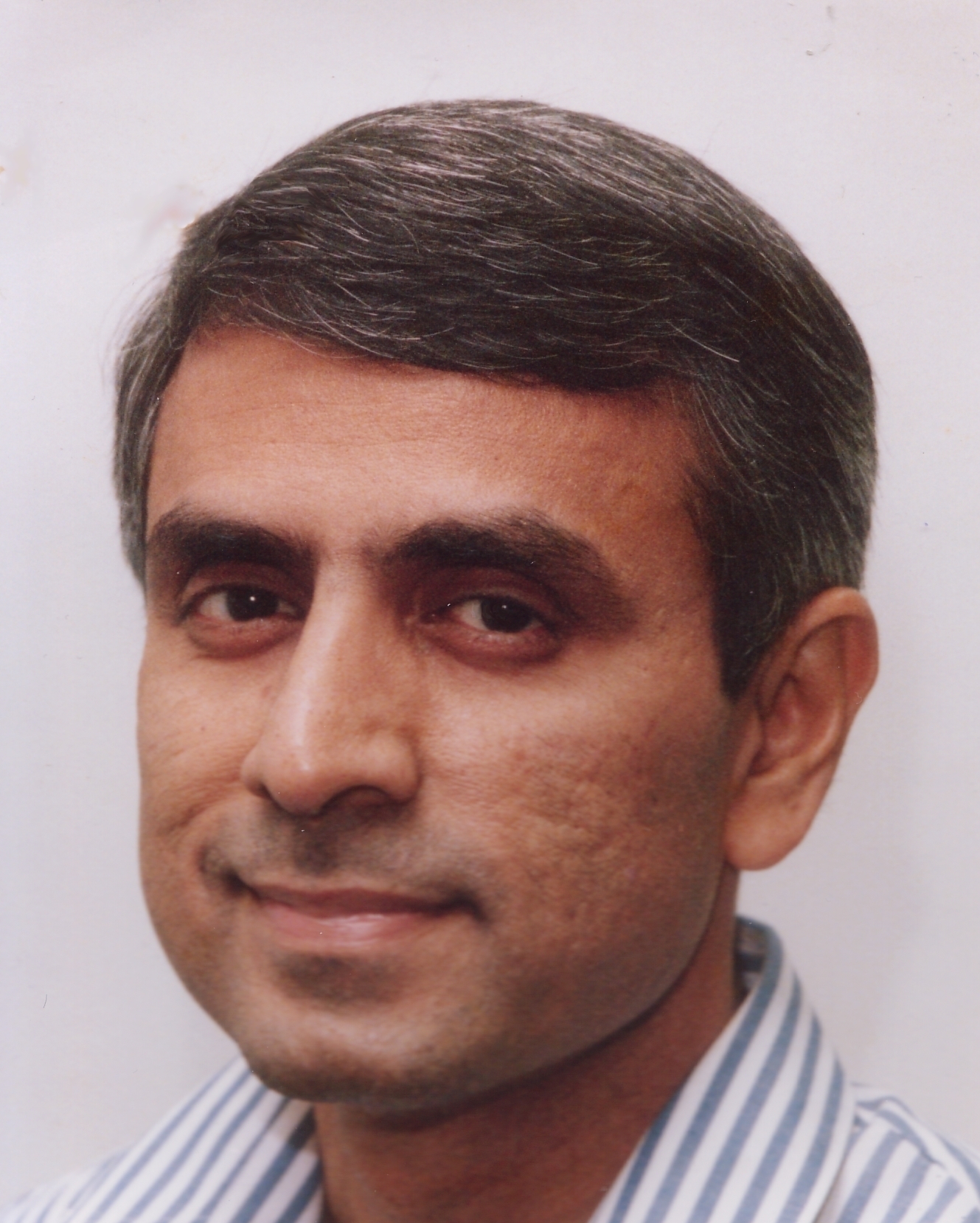}}]{Sridha Sridharan} has obtained an MSc (Communication Engineering) degree from the University of Manchester, UK, and a PhD degree from the University of New South Wales, Australia. He is currently with the Queensland University of Technology (QUT) where he is a Professor in the School of Electrical Engineering and  Robotics. He has published over 600 papers consisting of publications in journals and in refereed international conferences in the areas of Image and Speech technologies during the period 1990-2023.  During this period he has also graduated 85  PhD students in the areas of Image and Speech technologies. Prof Sridharan has also received a number of research grants from various funding bodies including the Commonwealth competitive funding schemes such as the Australian Research Council (ARC) and the National Security Science and Technology (NSST) unit. Several of his research outcomes have been commercialised.
\end{IEEEbiography}

\begin{IEEEbiography}[{\includegraphics[width=1in,height=1.25in,clip,keepaspectratio]{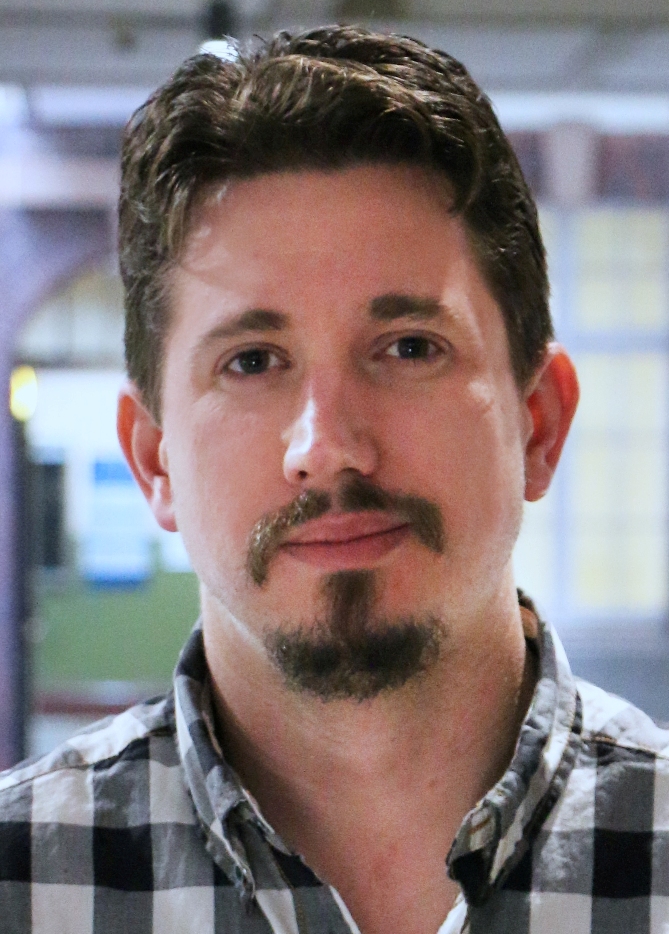}}]{Simon Denman} is an Associate Professor in the School of Electrical Engineering and Robotics at Queensland University of Technology (QUT). Simon actively researches in the fields of computer vision and machine learning, including action and event recognition, trajectory prediction, video analytics, biometrics, and medical signal processing. Simon has published over 200 papers in the areas of computer vision and machine learning, and co-leads the Applied Data Science research programme within the QUT Centre for Data Science.
\end{IEEEbiography}

\vfill

\end{document}